\documentclass{article} \usepackage{iclr2027_conference,times}

\usepackage{amsmath,amsfonts,bm}

\def\eqref#1{equation~\ref{#1}}

\def\1{\bm{1}}

\DeclareMathAlphabet{\mathsfit}{\encodingdefault}{\sfdefault}{m}{sl}
\SetMathAlphabet{\mathsfit}{bold}{\encodingdefault}{\sfdefault}{bx}{n}

\usepackage{hyperref}
\usepackage{url}
\usepackage{xspace}
\usepackage{wrapfig}
\usepackage{booktabs}
\usepackage{enumitem}
\usepackage{subcaption}
\usepackage{graphicx}
\definecolor{darkgreen}{rgb}{0.0, 0.5, 0.0}
\definecolor{darkred}{rgb}{0.6, 0.0, 0.0}
\definecolor{gray}{rgb}{0.5, 0.5, 0.5}
\definecolor{myblue}{RGB}{0, 0, 180}
\definecolor{mygreen}{RGB}{0, 108, 0}
\definecolor{myred}{RGB}{196, 0, 0}
\definecolor{darkmagenta}{RGB}{149, 0, 149}

\hypersetup{
    colorlinks=true,
    linkcolor=myred,
    citecolor=mygreen,
    filecolor=darkmagenta,
    urlcolor=darkmagenta
}

\newtheorem{theorem}{Theorem}

\newcommand{\MI}{\ensuremath{\mathsf{AMIGA}}\xspace}

\usepackage{tikz}
\usetikzlibrary{arrows.meta,positioning,fit}

\title{Graph Additive Models disentangling feature, topology, and interactions}

\author{Clara \'Alvarez-Rodr\'iguez\\
Universidad Carlos III de Madrid
\And
Steve Azzolin\\
University of Trento
\And
Vincenzo Marco De Luca\\
University of Trento
\AND
Filippo Maria Bianchi\\
The Arctic University of Norway
\And
Andrea Passerini\\
University of Trento
\And
Antonio Longa\thanks{Corresponding author: \texttt{antonio.longa@uit.no}}\\
The Arctic University of Norway
}

\begin{document}

\maketitle

\begin{abstract}

Graph Neural Additive Networks (GNANs) have emerged as an effective, interpretable-by-design graph architecture for learning on graphs. They achieve model transparency by combining per-feature shape functions with a simple distance-based reweighting. Despite the effectiveness of GNAN, we identify the following limitations:
\textit{i)} node-wise relevance scores are computed as a weighted aggregation over all graph nodes, allowing irrelevant nodes to inherit spurious relevance from relevant ones; 
\textit{ii)} graph topology is encoded only through shortest-path distances, and structurally distinct graphs may have identical distances;
\textit{iii)} feature and distance information are combined multiplicatively, thus entangling feature, distance, and interaction effects into a single, rescalable term. 
To address these limitations, we introduce \MI, which represents topology through a vocabulary of mined motifs and decomposes the contribution of each node into three additive channels: a feature channel, a structural channel, and a feature–structure interaction channel. We center these channels around a training-data baseline that gives every graph equal weight, regardless of its size, making the decomposition well defined and unique.
Across ten graph classification and regression benchmarks, \MI outperforms GNAN on most datasets and narrows the gap to standard black-box GNNs. Because every channel is centered and attached to a single node, the explanations of \MI stay local to the node that generates them and reconstruct the model output exactly while providing interpretability at different levels of granularity. Code is publicly available online\footnote{\url{https://anonymous.4open.science/r/AMIGA-2C36/}}.
\end{abstract}

\section{Introduction}
\label{sec:introduction}

Graph Neural Networks (GNNs) have emerged as a powerful tool for learning on graphs \citep{scarselli2008graph, bacciu2020gentle}, with applications in high-stakes scenarios like power grid analysis \citep{varbella2024powergraph}, health forecasting \citep{hu2024self}, software vulnerability detection \citep{lekeufack2024rapid}, and drug discovery \citep{wong2024discovery}.
However, the lack of transparency in black-box GNNs has motivated interpretable architectures that trade predictive performance for model transparency \citep{rudin2019stop}.

The prominent examples of these interpretable-by-design architectures are Graph Neural Additive Networks (GNANs) \citep{bechler2024intelligible}, which extend Generalized Additive Models \citep{hastie1986generalized} to graphs.
They learn one univariate function per feature, and linearly combine these over all nodes, scaling the contribution of each node by a learnable, univariate distance function.
This composition of simple, univariate functions gives users full knowledge of how the model aggregated information, and of which features are mostly relevant, enabling model-level understanding and debugging. GNANs have been further extended to temporal, heterogeneous data \citep{bechler2025superman}, to incorporate parameter-efficient transformations \citep{siddartha2026interpretable}, and to Graph Retrieval-Augmented Generation for LLMs \citep{sade2026ex}.

Despite the effectiveness of GNANs, in this work we identify the following three limitations that can hinder both their downstream performance and their interpretability:
\textit{i)} the node-wise importance scores extracted by GNAN for node $i$ do not represent its own contribution, but are a blurred aggregation of the importances of every node; \textit{ii)} GNAN relies exclusively on shortest-path distances to encode graph structure, potentially missing other informative topological properties.
\textit{iii)} GNAN multiplies the univariate feature function with the shortest-path distance reweighting, resulting in a less interpretable mixture of feature, distance, and interaction effects.

To overcome these limitations, we introduce the \emph{Attribute-Motif Interaction Graph Additive model} (\MI).
\MI relies on motifs automatically discovered from the training set, with each node contributing based on its features and on which motifs it appears in.
The model linearly combines three additive channels -- feature, structural, and feature–structure interaction -- each centered under a graph-balanced reference distribution that assigns equal mass to each training graph, ensuring the three channels form a unique canonical decomposition.
Summarizing, our main contributions are:
\begin{itemize}[leftmargin=1.25em]
        \item We identify three limitations of current graph additive models: node-wise importance scores do not isolate node-local contributions, shortest-path distances miss richer topology, and multiplicative feature–distance coupling obscures the separate feature, distance, and interaction effects.
        \item We introduce \MI, an interpretable-by-design GNN that replaces distance-based aggregation with automatically mined motifs, enabling the model to capture richer structural patterns while providing disentangled feature- and motif-based contributions.
        \item We develop a graph-balanced centering scheme that makes the additive decomposition well-defined and unique, allowing predictions to be decomposed into separate attribute, structural, and attribute–structure interaction contributions.
\end{itemize}

Across ten graph classification and regression benchmarks, \MI  outperforms GNAN on eight datasets and G-NAMRFF on nine.
Furthermore, in a controlled experiment, \MI successfully recovers the known task composition from label supervision alone, and achieves an explanation accuracy of $0.99$ and $1.00$ of AUCROC score on two synthetic datasets with known ground-truth explanations.
Although black-box GNNs remain stronger on most real-world benchmarks, \MI narrows the gap while providing an exact, interpretable decomposition of every prediction.

\section{Related works}

\textbf{Interpretable Graph Models.}
Interpretable-by-design GNNs were proposed to overcome the limitations of traditional post-hoc explainers \citep{rudin2019stop,longa2025explaining} by making the explanation part of the model's computation.
Examples include prototype-based GNNs -- predicting from the similarity between the input and learned prototypes \citep{zhang2022protgnn,ragno2022prototype} -- and self-explainable GNNs -- where a detector extracts a compact subgraph used by a classifier \citep{miao2022interpretable, azzolin2025beyond}.
Nonetheless, both families were shown to be potentially unfaithful \citep{christiansen2023faithful,azzolin2026gnn}. Recently, \citet{ragno2026logic} proposed to build GNN layers expressing the model's decisions as logical rules. This construction is, however, limited to classification tasks, and for complex tasks it may yield long and deeply nested formulas, hindering human understanding~\citep{miller1956magical}.
We focus on graph neural additive models \citep{bechler2024intelligible} because their visual explanations have the potential to be more easily interpretable.

\textbf{Motif- and Subgraph-Based Graph Representations.}
Recurring subgraphs, or motifs, capture higher-order structure in complex networks~\citep{milo2002network}. Standard message-passing GNNs are bounded by the 1-WL test and can therefore fail to distinguish graphs that differ in simple substructures, such as triangles~\citep{xu2018powerful,morris2019weisfeiler,chen2020can}. Subgraph- and motif-aware methods address this limitation by augmenting nodes with local substructure counts \citep{bouritsas2022improving}, processing collections of subgraphs~\citep{bevilacqua2022equivariant}, or constructing motif vocabularies for richer node and graph representations~\citep{yu2022molecular,wang2023motif,wang2025motif}.
Similarly to previous work integrating motif vocabularies, \MI achieves this whilst keeping the feature and interaction effects separate, enabling model transparency by design.

\section{Preliminaries}
\label{sec:preliminaries}

\subsection{Graph Neural Additive Models}
\label{sec:gnan}

Consider a graph \(G=(V,E)\), where each node \(j\in V\) has a feature vector \(x_j\in\mathbb{R}^{d}\). GNANs \citep{bechler2024intelligible} learn one univariate shape function $f_k: \mathbb{R} \mapsto \mathbb{R}$, usually implemented as an MLP, for each feature \(k\), together with a shared univariate function $\rho: \mathbb{R} \mapsto \mathbb{R}$ that controls how information is weighted across the graph.
Let \(\ell_{ij} = \operatorname{dist}(j,i)\) denote the shortest-path distance from a node \(j\) to a node \(i\), and let
\(
    n_i(\ell)
                            \)
be the number of nodes at distance \(\ell\) of node \(i\). GNAN assigns node \(i\) the feature-wise representation:
\begin{equation}
    [h_i]_k
    =
    \sum_{j\in V}
    \frac{1}{n_i(\ell_{ij})}
    \rho\!\left(\frac{1}{1+\ell_{ij}}\right)
    f_k(x_{jk}).
    \label{eq:gnan-node-representation}
\end{equation}

A later extension to this framework, namely G-NAMRFF~\citep{siddartha2026interpretable}, replaces GNAN’s all-pairs distance weighting with a polynomial graph filter \(H_G=\sum_{r=0}^{R}\alpha_r\widetilde{L}_G^r\). For each node \(i\), it applies an RFF-parameterized univariate response \(g_k\) to every filtered feature coordinate and computes \(h_i=\sum_{k=1}^{d}g_k((H_GX)_{ik})\). See \citet{siddartha2026interpretable} for details. For graph-level prediction, both models sum the node-wise representations into $\sum_{i\in V}\sum_{k=1}^{d}[h_i]_k$ followed by a task-specific activation function, like the sigmoid.

To explain a prediction, GNANs extract the importance score of node $i$ by summing Equation~\ref{eq:gnan-node-representation} over features and over every node in the graph, as follows:
\begin{equation}
    R_i^{\mathrm{GNAN}}
    =
    \sum_{j\in V}
    \frac{1}{n_i(\ell_{ij})}
    \rho\!\left(\frac{1}{1+\ell_{ij}}\right)
    \sum_{k=1}^{d} f_k(x_{jk}).
    \label{eq:gnan-receiver-attribution}
\end{equation}

\subsection{Limitations of Current Graph Neural Additive Models}
\label{sec:gnan-limitations}

\begin{wrapfigure}[22]{r}{0.4\columnwidth}
    \centering
    \begin{subfigure}{\linewidth}
        \centering
        \resizebox{1\linewidth}{!}{            \begin{tikzpicture}[
                node distance=18mm,
                every node/.style={font=\small},
                dot/.style={
                    circle,
                    draw,
                    minimum size=8mm,
                    inner sep=0
                }
            ]
                \node[dot, fill=blue!25] (B) {u};
                \node[dot, fill=red!15, right=of B] (R) {v};

                \draw[thick] (B) -- (R);

                \node[below=2mm of B]
                    {$\sum_k f_k(x_{uk})=1$};
                \node[below=8mm of B]
                    {$R_u^{\mathrm{GNAN}}=1$};
                                
                \node[below=2mm of R]
                    {$\sum_k f_k(x_{vk})=0$};
                \node[below=8mm of R]
                    {$R_v^{\mathrm{GNAN}}=0.5$};
                                            \end{tikzpicture}        }
        \caption{GNAN's local explanation.}
        \label{fig:limitation2}
    \end{subfigure}

    \vspace{0.6em}

    \begin{subfigure}{\linewidth}
        \centering
        \includegraphics[
            width=0.55\linewidth
        ]{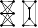}
        \caption{\(K_{3,3}\) and the triangular prism.}
        \label{fig:limitation1}
    \end{subfigure}
\caption{\textbf{Two limitations of GNAN}: (a) node relevance mixes neighboring feature contributions; (b) shortest-path profiles cannot distinguish \(K_{3,3}\) from the triangular prism.}
                                \label{fig:limitations-combined}
\end{wrapfigure}

Both GNAN and G-NAMRFF share these limitations:

\textbf{Node-wise relevance scores are blurred.}
GNAN computes the relevance score of node $i$ by aggregating feature contributions from all nodes \(j\), weighted by their distance from \(i\), see Equation~\ref{eq:gnan-receiver-attribution}.
Consequently, this score does not represent node \(i\)’s own feature contribution, but in conjunction with every other node in the graph.
For example, a node with zero feature contribution on its own can receive a non-zero relevance score just because it is close to a node with a positive score. 
Figure~\ref{fig:limitation2} illustrates this case: node \(u\) carries positive contribution \(\sum_k f_k(x_{uk})=1\), whilst node \(v\) carries none \(\sum_k f_k(x_{vk})=0\).
With a distance weighting $\rho$ of one from a node to itself and \(0.5\) between \(u\) and \(v\), Equation~\ref{eq:gnan-receiver-attribution} gives \(R_u^{\mathrm{GNAN}}=1\) and \(R_v^{\mathrm{GNAN}}=0.5\),
even though the only nonzero feature contribution originates from \(u\). G-NAMRFF faces a similar issue, as it mixes neighboring features within its finite receptive field before assigning relevance scores.

\textbf{GNAN and G-NAMRFF can miss relevant topological information.}
\label{sec:gnan-distance-limitation}
GNAN represents topology through shortest-path distance.
This can identify how far nodes lie from one another, but it does not determine which subgraphs connect them. 
For instance, Figure~\ref{fig:limitation1} shows an example of the graph \(K_{3,3}\) and the triangular prism, where both have six nodes with the same distance profile from every node.
Hence, Equation~\ref{eq:gnan-node-representation} scores them identically for every \(\rho\) and \(\{f_k\}_{k=1}^d\), even though only the prism contains a triangle.
Laplacian filtering does not resolve this limitation in general: with constant node features, both graphs are 3-regular, so $\widetilde L_G\mathbf{1}=0$ and G-NAMRFF also maps them identically.

\textbf{Features, topology, and interaction effects are mixed.}
\label{sec:gnan-decomposition-limitation}
GNAN combines the feature and distance function multiplicatively.
This product, however, does not separate the effect of features from that of topological information, and how these interact.
Consider the following example, with $f(x)=3+x$ and $\rho(u)=2+u$.
Then, $\rho(u)f(x)$ yields $6+2x+3u+xu$ which is composed of a constant term, a feature-only term, a distance-only term, and a joint term.
Crucially, all these terms are implicit in the multiplication.
In addition, the individual factors also lack a fixed scale, such that for any nonzero \(\gamma\),
\(
    \rho(u)f(x)
    =
    \bigl[\gamma f(x)\bigr]
    \bigl[\rho(u)/\gamma\bigr].
\)
This implies that, although the two univariate functions and hence the explanations change, the prediction remains the same.
Interpreting their absolute magnitudes is, therefore, difficult and requires an explicit reference convention.
G-NAMRFF replaces the GNAN product with $g_k((H_GX)_{ik})$, but features and topology have already been mixed inside its argument, and therefore the same argument above applies.

\begin{figure}[t]
    \centering
    \includegraphics[width=1\linewidth]{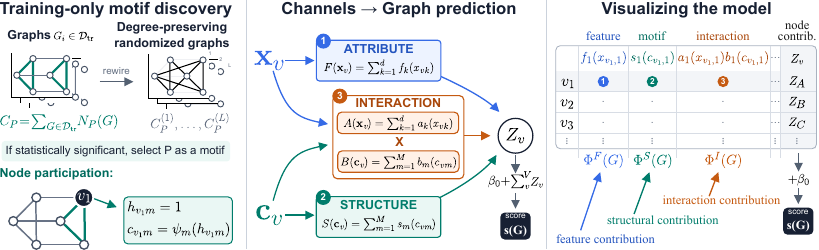}
    \caption{\textbf{Overview of \MI}. \textbf{Left:} motifs are selected when they occur significantly more often than in randomly rewired training graphs with unchanged node degrees. Each node is then described by its participation in the selected motifs. \textbf{Center:} each node contribution is decomposed into feature, structural, and feature--motif interaction channels, and then pooled into the graph score.
                    \textbf{Right:} this decomposition represents the model-level explanation of \MI, and enables full model transparency by independently inspecting feature, structural, and interaction contributions, allowing the model's computations to be exactly reconstructed.
}
\label{fig:overview}
\end{figure}

\section{\MI: Attribute-Motif Interaction Graph Additive Model}
\label{sec:amiga}

Figure~\ref{fig:overview} summarizes \MI, which replaces distance-based structural summaries with a motif vocabulary and decomposes each node contribution into feature, structural, and interaction effects. Using the training set $\mathcal{D}_{\mathrm{tr}}$ 
only, \MI selects induced motifs that occur significantly more often than in randomized training graphs that preserve the degree of every node. The resulting vocabulary $\mathcal{P}$ is fixed and used unchanged for validation and test graphs. Each node $v$ is represented by its attributes $\mathbf{x}_v$ and its normalized motif-membership counts $\mathbf{c}_v$. \MI then computes
\begin{equation}
z_v
= F(\mathbf{x}_v)
+ S(\mathbf{c}_v)
+ \underbrace{A(\mathbf{x}_v)B(\mathbf{c}_v)}
  _{I(\mathbf{x}_v,\mathbf{c}_v)},
\qquad
s(G)=\beta_0+\sum_{v\in V} z_v .
\label{eq:node-contribution}
\end{equation}
The three terms capture feature, structural, and feature--motif interaction contributions. Their shape functions are centered under a graph-balanced training reference, yielding node-local contributions that reconstruct $s(G)$ exactly.

\subsection{Motif discovery and node participation}
\label{sec:motif-representation}

Using $\mathcal{D}_{\mathrm{tr}}$ only, we enumerate connected, unlabeled graphlets with $q\in\{3,\ldots,q_{\max}\}$ and count their induced occurrences. We retain candidates whose training-set counts are enriched relative to $L$ degree-preserving rewired replicas after
Benjamini--Hochberg correction at level $\alpha$. The resulting vocabulary $\mathcal{P}=\{P_1,\ldots,P_M\}$ is frozen for validation and test graphs. Appendix~\ref{sec:motif-discovery} gives the empirical test, enumeration settings, vocabulary sizes, and runtimes.

For $P_m\in\mathcal{P}$, let $\mathcal{O}_m(G)$ denote its induced
occurrences and let $h_{vm}=|\{U\in\mathcal{O}_m(G):v\in U\}|$.
We use $c_{vm}=[\log(1+h_{vm})-\widehat{\mu}_m]/\widehat{\sigma}_m$,
with $\widehat{\mu}_m$ and $\widehat{\sigma}_m$ estimated from the training nodes. The logarithm compresses large counts, while standardization aligns motif scales. Raw counts and occurrences are retained to localize the resulting explanations.

\subsection{Feature, structural, and interaction channels}
\label{sec:model-channels}

Each channel is built from centered univariate shape functions, defined precisely in Section~\ref{sec:canonicalization}:
\[
    F(x)
    =
    \sum_{k=1}^{d} f_k(x_k),
    \quad
    S(c)
    =
    \sum_{m=1}^{M} s_m(c_m),
    \quad
    A(x)
    =
    \sum_{k=1}^{d} a_k(x_k),
    \quad
    B(c)
    =
    \sum_{m=1}^{M} b_m(c_m). 
    \] 

The interaction therefore expands as \(I(x,c) := A(x)B(c)
= \sum_{k=1}^{d}\sum_{m=1}^{M} a_k(x_k)b_m(c_m)\). Accordingly, the node-level interaction contribution is \(I_v := I(x_v,c_v)\). Every interaction term involves one feature and one motif. The model learns $d+M$ interaction shapes, while exposing all $dM$ feature--motif pairs. Although the $dM$ pairwise functions could instead be learned directly, we favor the factorized form since humans can only track a few items at once \citep{cowan2001magical}. This keeps the number of independently inspectable functions traceable as $d$ and $M$ grow.
The uncentered functions $\widetilde f_k$, $\widetilde s_m$, $\widetilde a_k$, and $\widetilde b_m$ are parameterized by small univariate MLPs for numerical features and motif coordinates. For categorical features, $\widetilde f_k$ and $\widetilde a_k$ are scalar lookup tables with one learned value per category; when a categorical variable is supplied as a one-hot vector, the full group is treated as a single variable. The centered shape functions used above are obtained from these uncentered functions as described in Section~\ref{sec:canonicalization}, and each can be inspected directly.

\subsection{Graph-balanced canonicalization}
\label{sec:canonicalization}

Equation~\ref{eq:node-contribution} separates the node score into feature, structural, and interaction channels. This decomposition is not identifiable because constants can be shifted between the channels and the intercept without changing the represented function. We remove this ambiguity by centering the channels. Centering, however, requires choosing a reference distribution. A simple average over all training nodes would give larger graphs greater weight. We therefore define a graph-balanced reference distribution by averaging first within each graph and then across graphs:

\[
\mathbb{E}^{\mathrm{gb}}_X[u(X)]
=
\frac{1}{|\mathcal D_{\mathrm{tr}}|}
\sum_{G\in\mathcal D_{\mathrm{tr}}}
\frac{1}{|V(G)|}
\sum_{v\in V(G)} u(x_v),
\;
\mathbb{E}^{\mathrm{gb}}_C[w(C)]
=
\frac{1}{|\mathcal D_{\mathrm{tr}}|}
\sum_{G\in\mathcal D_{\mathrm{tr}}}
\frac{1}{|V(G)|}
\sum_{v\in V(G)} w(c_v).
\]
These averages define the graph-balanced empirical distributions
$\mu_X^{\mathrm{gb}}$ and $\mu_C^{\mathrm{gb}}$ of features and motif
participation, respectively. We center every feature shape under $\mu_X^{\mathrm{gb}}$ and every motif shape under $\mu_C^{\mathrm{gb}}$. For example,
\[
f_k(t)
=
\widetilde f_k(t)
-
\mathbb{E}^{\mathrm{gb}}_X[\widetilde f_k(X_k)],
\qquad
s_m(t)
=
\widetilde s_m(t)
-
\mathbb{E}^{\mathrm{gb}}_C[\widetilde s_m(C_m)],
\]
and analogously for $a_k$ and $b_m$. Hence
\(
\mathbb{E}^{\mathrm{gb}}_X[F(X)]
=
\mathbb{E}^{\mathrm{gb}}_X[A(X)]
=
\mathbb{E}^{\mathrm{gb}}_C[S(C)]
=
\mathbb{E}^{\mathrm{gb}}_C[B(C)]
=
0.
\)
For categorical variables, these expectations use their graph-balanced
empirical frequencies.

Let \(\mu = \mu_X^{\mathrm{gb}} \otimes \mu_C^{\mathrm{gb}}\) be the product of the graph-balanced marginals. Under $\mu$, $X \sim \mu_X^{\mathrm{gb}}$ and $C \sim \mu_C^{\mathrm{gb}}$ are sampled
independently. We use $\mu$ as the reference distribution for centering,
which makes the separation into feature, structural, and interaction
components unique. This choice does not assume that $X$ and $C$ are
independent in the observed data. Their empirical joint distribution may
contain dependencies and may exclude some pairs $(x,c)$ included in the
product support.

\begin{theorem}[Canonical modality decomposition]
\label{thm:anova}
Let
$\mathcal X_{\mathrm{tr}}=\operatorname{supp}(\mu_X^{\mathrm{gb}})$ and
$\mathcal C_{\mathrm{tr}}=\operatorname{supp}(\mu_C^{\mathrm{gb}})$.
For any function
$g:\mathcal X_{\mathrm{tr}}\times\mathcal C_{\mathrm{tr}}\to\mathbb R$,
there exist unique functions $g_X$, $g_C$, and $g_{XC}$ and a unique
constant $g_0$ such that, for every
$(x,c)\in\mathcal X_{\mathrm{tr}}\times\mathcal C_{\mathrm{tr}}$,
\begin{equation}
g(x,c)
=
g_0+g_X(x)+g_C(c)+g_{XC}(x,c),
\label{eq:canonical-decomposition}
\end{equation}
with 
\(\mathbb{E}^{\mathrm{gb}}_X[g_X(X)]
=
\mathbb{E}^{\mathrm{gb}}_C[g_C(C)]
= \mathbb{E}^{\mathrm{gb}}_X[g_{XC}(X,c)]=\mathbb{E}^{\mathrm{gb}}_C[g_{XC}(x,C)]=0\), \(\forall c\in\mathcal C_{\mathrm{tr}}\), \(\forall x\in\mathcal X_{\mathrm{tr}}\).

The four terms are pairwise orthogonal under the inner product
$\langle p,q\rangle_\mu=\mathbb{E}_\mu[p(X,C)q(X,C)]$.

For $H(x,c)=A(x)B(c)$, the canonical interaction is
\[
H_{XC}(x,c)
=
\bigl(A(x)-\mathbb{E}^{\mathrm{gb}}_X[A(X)]\bigr)
\bigl(B(c)-\mathbb{E}^{\mathrm{gb}}_C[B(C)]\bigr).
\]
\end{theorem}

The proof is given in Appendix~\ref{app:proof-anova}. For the centered model,
Theorem~\ref{thm:anova} yields
$g_0=0$, $g_X=F$, $g_C=S$, and $g_{XC}=AB$.
Hence, the three modality-level channels $F$, $S$, and $AB$ are uniquely determined on the empirical product support. This uniqueness applies to the channels as whole functions. It does not extend to their coordinate-wise representations through $f_k$, $s_m$, $a_k$, and $b_m$, since different collections of shape functions may produce the same modality-level channels. Therefore, each term $a_k(x_k)b_m(c_m)$ exactly describes the contribution produced by the fitted parameterization, although an equivalent parameterization may distribute the same interaction differently across feature--motif pairs.

To promote sparse explanations and choose a balanced magnitude within the remaining reciprocal-rescaling class, we minimize $\mathcal{L}=\mathcal{L}_{\mathrm{task}}+\lambda\Omega$, where $\Omega$ sums the graph-balanced mean absolute values of the centered shape responses and $\lambda\geq 0$. Along the rescaling $A\mapsto\gamma A$, $B\mapsto B/\gamma$, the interaction and every feature--motif contribution remain unchanged, while the equal penalties on the two factors favor a balanced scale; Appendix~\ref{app:sparsity-scale} gives the full penalty and calculation.

\subsection{Factorized Motif Explanations}
\label{sec:fme}

Factorized Motif Explanations (FME) read the quantities already computed
by the model. No surrogate model or post-hoc approximation is required.

\textbf{Graph and node explanations.}
The three graph-level channel totals are
\begin{equation}
    \Phi^F(G)
    =
    \sum_{v\in V}F(x_v),
    \qquad
    \Phi^S(G)
    =
    \sum_{v\in V}S(c_v),
    \qquad
    \Phi^I(G)
    =
    \sum_{v\in V}A(x_v)B(c_v),
    \label{eq:channel-totals}
\end{equation}
which reconstruct the score, \(s(G)=\beta_0+\Phi^F(G)+\Phi^S(G)+\Phi^I(G)\).
At node level, FME reports the three terms \(F(x_v)\), \(S(c_v)\),
\(A(x_v)B(c_v)\), or their sum \(z(x_v,c_v)\).

\textbf{Motif and feature--motif explanations.}
The contribution of motif type $m$ to the structural channel, and of
feature--motif pair $(k,m)$ to the interaction, are
\[
    Q^S_m(G)
    =
    \sum_{v\in V}s_m(c_{vm}),
    \qquad
    Q^I_{km}(G)
    =
    \sum_{v\in V}
    a_k(x_{vk})b_m(c_{vm}) \]
with \(\Phi^S(G)=\sum_{m=1}^{M}Q^S_m(G)\) and
\(\Phi^I(G)=\sum_{k=1}^{d}\sum_{m=1}^{M}Q^I_{km}(G)\). The first identity
ranks reusable motif templates; the second produces a \(d\times M\) matrix
of feature--motif contributions.

\textbf{Exactness and occurrence explanations.}
The identities above exactly reconstruct the graph score and the structural and interaction channels. For each node and motif type, FME computes the change in the structural and interaction contributions of the node when motif participation moves from zero to the observed level. It then distributes this change equally across all occurrences of the motif that contain the node. Adding the zero-participation baseline terms recovers the original motif contributions. These allocations remain valid when motif occurrences overlap and are equivariant to vertex relabeling. Their definitions and verification are given in Appendix~\ref{app:occurrence-allocation}. They describe the model computation and should not be interpreted as the causal effects of deleting a motif.

\section{Experimental Evaluation}
\label{sec:experiments}

We evaluate \MI along three axes: whether it recovers known feature, structural, and interaction effects; whether its explanations identify
ground-truth predictive patterns; and whether its transparency preserves
useful predictive performance.

\textbf{Experimental Protocol.}
We evaluate four graph-classification datasets
(Mutagenicity, NCI1, PROTEINS, and PTC-MR) and six graph-regression datasets
(BA-Reg~1, BA-Reg~2, Crippen, QM9-$\alpha$, QM9-HOMO, and QM9-$\mu$).
We compare \MI with standard, widely used black-box GNNs like GCN \citep{kipf2016semi}, GIN \citep{xu2018powerful}, and Graph Transformer \citep{shi2020masked}, and with the interpretable GNAN and G-NAMRFF baselines under the same splits and preprocessing. 
For each seed, motif discovery, normalization, and centering use only the corresponding training split. Checkpoints are selected on validation performance. Results 
are mean $\pm$ standard deviation over five seeds, using AUROC for
classification and MAE for regression.
Dataset statistics and preprocessing are reported in Appendix~\ref{app:datasets}; optimization, hyperparameter selection, and training details are given in Appendix~\ref{app:training}.

\subsection{Recovering Known Effects}
\label{sec:synthetic_recovery}

\begin{figure}[t]
    \centering
    \includegraphics[width=\linewidth]
    {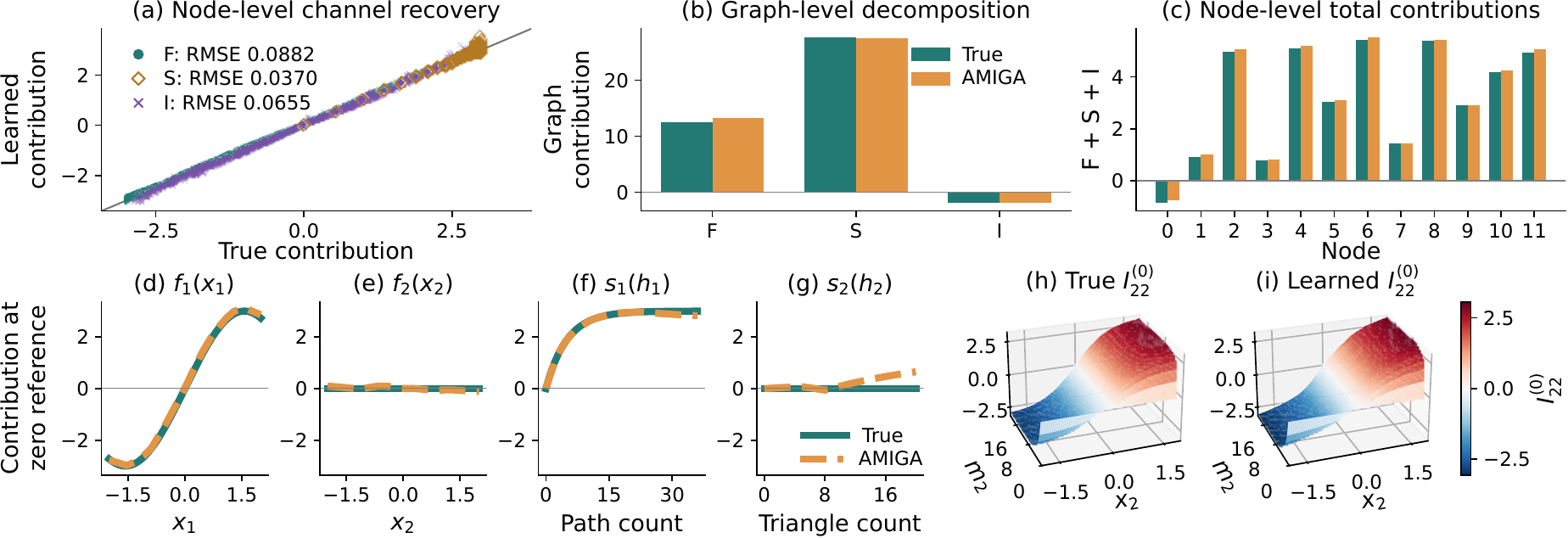}
    \caption{
                \textbf{\MI can recover known effects from only label supervision.}
                \textbf{(a)} True and learned node-level channel contributions.
        \textbf{(b)} Graph-level channel totals for one test graph.
        \textbf{(c)} Node-level total contributions for the same graph.
        \textbf{(d--g)} Active and inactive main effects.
        \textbf{(h--i)} True and learned $I_{22}^{(0)}$, shown with common
        axes and color scale. 
    }
    \label{fig:toy-recovery}
\end{figure}

We first test whether graph-level supervision suffices to recover a known
node-level decomposition. 
We generate 2,000 Erdős--Rényi graphs with 12
nodes and edge probability sampled uniformly from $[0.1,0.4]$. Node features
$x_{v1},x_{v2}\sim\mathcal{U}[-2,2]$ are independent of the graph topology.
Let $h_{v1}$ and $h_{v2}$ denote the numbers of induced three-node paths and
triangles containing node $v$. The target is
\begin{equation*}
y(G)=\sum_{v\in V}\left[
3\sin(x_{v1})
+3(1-e^{-h_{v1}/5})
+3\tanh(x_{v2})(1-e^{-h_{v2}/2})
\right]+\epsilon_G,
\end{equation*}
where $\epsilon_G\sim\mathcal{N}(0,\sigma^2)$ with $\sigma$ equals 10\% of
the standard deviation of the training targets. We fix the motif
vocabulary to three-node paths and triangles, isolating effect recovery from
motif discovery. 

For comparison with the ground truth, we exactly re-anchor the learned
components at $x=0$ and $h=0$, using their resulting uncentered forms, leaving every prediction unchanged. 
Because all graphs have 12 nodes, the resulting per-node constant can be absorbed
into the graph intercept. Under this reference, only $f_1$, $s_1$, and
$I_{22}^{(0)}$ are active; $f_2$, $s_2$, and all other feature--motif interactions are zero. We report in Appendix~\ref{app:synthetic_details} the experimental details and prediction-preserving zero-reference transformation used in this analysis.

\MI obtains a test MAE of $1.02\pm0.42$, and Figure~\ref{fig:toy-recovery} reports a visualization example.
Over test nodes, the RMSEs of the recovered feature, structural, and interaction
contributions are $0.088$, $0.037$, and $0.065$, respectively.
The learned shapes recover the sinusoidal feature effect, the saturating path
effect, and the sign-changing triangle interaction, while the inactive
effects remain near zero.

\subsection{Explanation Plausibility}
\label{sec:explanation_plausibility}

\begin{wraptable}{r}{0.45\textwidth}
\centering
\scriptsize
\setlength{\tabcolsep}{4pt}
\caption{Explanation AUROC on BA-Reg~1 and BA-Reg~2, averaged over 5 seeds.}
\begin{tabular}{lcc}
\toprule
\textbf{Model} & \textbf{BA-Reg 1} & \textbf{BA-Reg 2} \\
\midrule
GIN + GNNExpl. & $0.94 \pm 0.11$ & $0.72 \pm 0.19$ \\
GIN + IG       & $0.88 \pm 0.11$ & $0.22 \pm 0.11$ \\
\midrule
GNAN           & $0.71 \pm 0.12$ & $0.45 \pm 0.16$ \\
G-NAMRFF       & $0.50 \pm 0.11$ & $0.54 \pm 0.08$ \\
\textbf{\MI} (ours)   & $\mathbf{0.99 \pm 0.02}$
               & $\mathbf{1.00 \pm 0.00}$ \\
\bottomrule
\end{tabular}
\label{tab:bareg-auroc}
\end{wraptable}

Faithfulness follows directly from the additive identities in Section~\ref{sec:fme}, as the reported contributions reconstruct the model score exactly.
We additionally measure \emph{plausibility}, i.e., how well the explanation recovers a ``ground-truth'' pattern \citep{longa2025explaining}.
This measures the agreement between the provided node-wise explanations and the data ground truth, measured via the AUROC metric.
This evaluation is restricted to BA-Reg~1 and BA-Reg~2, since they are the only datasets with annotated ground truth explanations.
The node-scoring rules, post-hoc baseline settings, and per-graph AUROC aggregation are detailed in Appendix~\ref{app:explanation_evaluation}.
As shown in Table~\ref{tab:bareg-auroc}, \MI achieves an AUROC of $0.99\pm0.02$ on BA-Reg~1 and $1.00\pm0.00$ on BA-Reg~2, outperforming both interpretable baselines and the post-hoc GNNExplainer \citep{ying2019gnnexplainer} and Integrated Gradients \citep{sundararajan2017axiomatic} methods on both tasks.

\subsection{Predictive Performance}
\label{sec:predictive_performance}

Table~\ref{tab:predictive_performance} reports predictive performance
across the ten benchmarks. \MI improves on GNAN on eight of the ten benchmarks
and on G-NAMRFF on nine. It outperforms both interpretable baselines on three
of four classification datasets and obtains the lowest overall MAE on BA-Reg~1 and BA-Reg~2. Black-box GNNs remain stronger on most real-world tasks, yet \MI narrows the performance gap in eight of the ten datasets while retaining an exact decomposition of every prediction into feature, structural, and interaction contributions. Training-time results in Appendix~\ref{app:training} show that \MI has the lowest runtime on nine of ten datasets. Channel ablations in Appendix~\ref{app:additional_experiments} show that the full \(F+S+I\) model is best on nine of ten datasets and isolate complementary roles for the interaction channel on BA-Reg~1 and the structural channel on BA-Reg~2.

\begin{table*}[ht]
\centering
\caption{
Predictive performance averaged over five seeds. Classification rows report test AUROC ($\uparrow$); regression
rows report test MAE ($\downarrow$). Bold marks the best result within each model group.
}
\label{tab:predictive_performance}
\setlength{\tabcolsep}{2.8pt}
\renewcommand{\arraystretch}{1.02}
\resizebox{\textwidth}{!}{
\begin{tabular}{@{}lccc|ccc@{}}
\toprule
&
\multicolumn{3}{c}{\textbf{Black-box models}} &
\multicolumn{3}{c}{\textbf{Interpretable models}} \\
\cmidrule(lr){2-4}
\cmidrule(lr){5-7}
\textbf{Dataset}
& \textbf{GCN}
& \textbf{GIN}
& \textbf{GTrans}
& \textbf{GNAN}
& \textbf{G-NAMRFF}
& \textbf{\MI} (ours) \\
\midrule

\multicolumn{7}{@{}l}{\textit{Graph classification: AUROC $\uparrow$}} \\

Mutagenicity
& $0.851 \pm 0.005$
& $\mathbf{0.868 \pm 0.018}$
& $0.849 \pm 0.010$
& $0.751 \pm 0.002$
& $0.698 \pm 0.068$
& $\mathbf{0.828 \pm 0.015}$ \\

NCI1
& $0.840 \pm 0.017$
& $\mathbf{0.868 \pm 0.013}$
& $0.807 \pm 0.007$
& $0.651 \pm 0.045$
& $0.734 \pm 0.042$
& $\mathbf{0.768 \pm 0.020}$ \\

PROTEINS
& $0.791 \pm 0.010$
& $\mathbf{0.811 \pm 0.012}$
& $0.697 \pm 0.020$
& $\mathbf{0.821 \pm 0.008}$
& $0.786 \pm 0.036$
& $0.733 \pm 0.046$ \\

PTC-MR
& $0.524 \pm 0.037$
& $0.518 \pm 0.044$
& $\mathbf{0.593 \pm 0.056}$
& $0.501 \pm 0.029$
& $0.555 \pm 0.056$
& $\mathbf{0.574 \pm 0.026}$ \\

\midrule
\multicolumn{7}{@{}l}{\textit{Graph regression: MAE $\downarrow$}} \\

BA-Reg 1
& $231.6 \pm 57.9$
& $\mathbf{108.4 \pm 23.8}$
& $327.0 \pm 8.4$
& $372.1 \pm 16.5$
& $441.2 \pm 16.3$
& $\mathbf{4.2 \pm 1.2}$ \\

BA-Reg 2
& $0.43 \pm 0.03$
& $\mathbf{0.17 \pm 0.14}$
& $4.91 \pm 0.00$
& $8.87 \pm 8.83$
& $4.91 \pm 0.00$
& $\mathbf{0.01 \pm 0.01}$ \\

Crippen
& $0.28 \pm 0.03$
& $\mathbf{0.18 \pm 0.03}$
& $0.62 \pm 0.06$
& $0.88 \pm 0.10$
& $1.20 \pm 0.22$
& $\mathbf{0.46 \pm 0.18}$ \\

QM9-$\alpha$
& $0.67 \pm 0.02$
& $\mathbf{0.44 \pm 0.01}$
& $2.14 \pm 0.10$
& $\mathbf{2.66 \pm 0.10}$
& $3.46 \pm 0.13$
& $3.35 \pm 0.03$ \\

QM9-HOMO
& $0.16 \pm 0.00$
& $\mathbf{0.12 \pm 0.00}$
& $0.14 \pm 0.00$
& $0.35 \pm 0.04$
& $0.33 \pm 0.03$
& $\mathbf{0.24 \pm 0.01}$ \\

QM9-$\mu$
& $0.64 \pm 0.00$
& $\mathbf{0.53 \pm 0.01}$
& $0.56 \pm 0.00$
& $0.87 \pm 0.01$
& $0.91 \pm 0.03$
& $\mathbf{0.85 \pm 0.01}$ \\

\bottomrule
\end{tabular}
}
\end{table*}

\section{From Global Rules to a Local Explanation}
\label{sec:mutagenicity_explanations}

Next, we showcase an example of how \MI provides explanations for the Mutagenicity dataset. Explanations are read directly from the model, without a surrogate or post-hoc approximation.
Because \MI channels are centered using graph-balanced training marginals, zero defines a common reference that gives every training graph equal weight, regardless of its size; positive and negative contributions respectively raise and lower the model score relative to this reference.

\begin{figure}[ht]
    \centering
    \includegraphics[width=\linewidth]
    {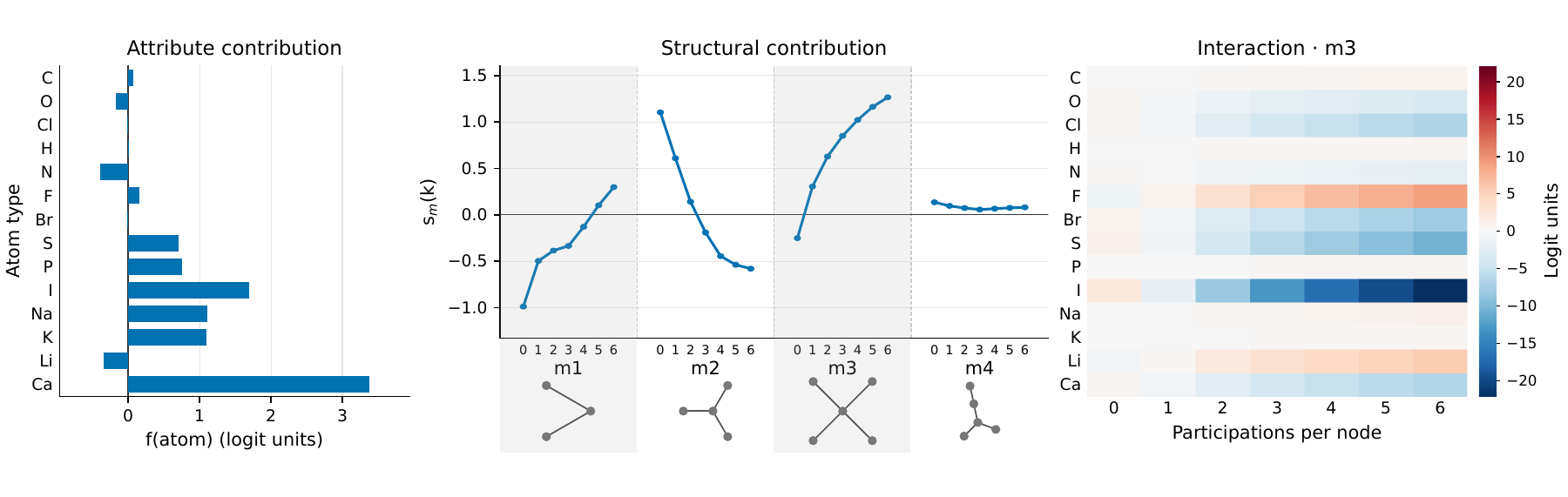}
    \caption{
        \textbf{Global response functions learned by \MI on Mutagenicity.}
        \textbf{Left:} centered atom-type effects $f(\mathrm{atom})$.
        \textbf{Center:} structural effects $s_m(k)$ against the number
        $k$ of motif occurrences containing a node.
        \textbf{Right:} interaction
        $I_3(\mathrm{atom},k)$ between atom identity and motif $m_3$.
        Positive and negative values raise and lower the positive-class
        logit relative to the graph-balanced reference.
    }
    \label{fig:global_explanation}
\end{figure}

\textbf{Global Decision Rules.}
Figure~\ref{fig:global_explanation} reveals a sparse, context-dependent decision rule. Most nonzero feature and structural effects are concentrated in a small set of atom types and motifs. For motif $m_3$, the structural effect $s_3(k)$ increases with node participation $k$. Its overall motif-related contribution, however, also depends on the participating atom, because the interaction $I_3(\mathrm{atom},k)$ changes both magnitude and sign across atom types. As participation increases, the interaction becomes positive for fluorine and negative for iodine, bromine, and sulfur. The same motif can therefore raise or lower the model score depending on the atom in which it occurs. The feature-only and structure-only channels capture the corresponding main effects, while the interaction channel additionally captures how the effect of motif participation varies across atom types.

\begin{figure}[ht]
    \centering
    \includegraphics[width=0.9\linewidth]
    {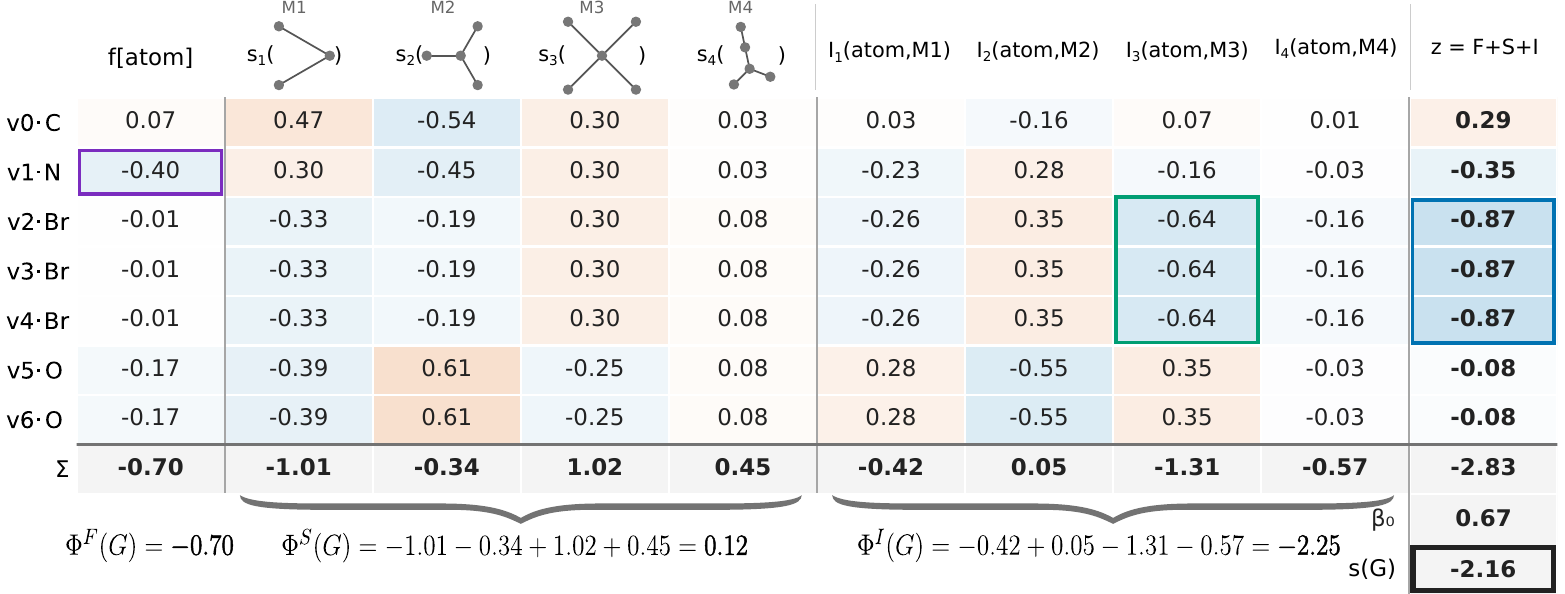}
    \caption{
        \textbf{Example of the exact \MI decomposition for one molecular graph}. Rows correspond to atoms; columns report the feature contribution, motif contributions, atom–motif interaction contributions, and node total $z_v$. The bottom row aggregates each component across nodes. Adding the bias $\beta_0$ reconstructs $s(G)$ exactly. Outlined cells mark the contributions discussed in the text.    }
    \label{fig:expla_mutag}
\end{figure}

\textbf{Local Explanations.}
Figure~\ref{fig:expla_mutag} contextualizes the global rule above over a molecule. Each row decomposes an atom contribution into feature, structural, and interaction effects. The nitrogen node $v_1$ contributes $-0.40$ through its atom type. The dominant repeated signal is the bromine $m_3$ interaction identified above: it contributes $-0.64$ to
each bromine node $v_2$, $v_3$, and $v_4$, driving each node total to
$-0.87$. At graph level, the feature channel contributes $\Phi^F(G)=-0.70$, while
the structural contributions largely cancel, yielding $\Phi^S(G)=0.12$. The interaction
channel dominates with $\Phi^I(G)=-2.25$. Hence
$\sum_v z_v=-2.83$, and adding the bias term $\beta_0=0.67$ gives the exact graph score $s(G)=-2.16$.
To conclude, this example shows how to contextualize the global decision rule above into a local interpretation of each node's relevance, for a given molecular graph.

\section{Conclusion}
\label{sec:conclusion}
We introduced \MI, an interpretable-by-design graph model that replaces distance-only summaries with mined motifs. Graph-balanced centering yields canonical feature, structural, and interaction channels, while Factorized Motif Explanations reconstruct predictions exactly from graph to feature--motif level. Across ten benchmarks, \MI outperformed GNAN and G-NAMRFF on eight and nine datasets, respectively. It also recovered known generative effects from graph-level supervision and identified ground-truth explanatory nodes on two synthetic tasks. These results show that richer structural modeling can coexist with exact model transparency.

\textbf{Limitations.} These results open several directions for extending \MI. Enriching the motif vocabulary and exploring more expressive, explicitly decomposable interactions may help close the remaining accuracy gap with black-box GNNs on several real-world benchmarks. Scalable counting algorithms or sampling-based motif estimates could extend the vocabulary beyond the small induced graphlets supported by the current exact enumeration procedure. Finally, Factorized Motif Explanations provide a faithful decomposition of the learned score and a starting point for formulating hypotheses about motif effects. Testing these hypotheses through controlled graph interventions, with explicit causal assumptions, is a promising direction toward causal analysis; the current allocations themselves do not estimate causal effects.

\clearpage

\subsection*{AI use statement}
In this work, we used OpenAI Codex with the GPT-5.6 Sol model. Within the research workflow, Codex assisted with interpreting model outputs and presenting the corresponding model explanations. It also helped draft portions of the manuscript, edit the text for clarity and readability, and prove mathematical claims by drafting portions of the proofs presented in Appendix~\ref{app:proofs}. The research questions, hypotheses, theoretical framework, mathematical claims, methodology, and experiments were developed by the authors.

The authors independently reviewed and verified all AI-assisted interpretations against the underlying model outputs, checked every step of each AI-assisted proof, and revised the final explanations and arguments. We did not use generative AI to implement methods, generate synthetic data, clean or reformat datasets, translate content, run experiments, or produce the reported experimental results. Generative AI performed no analysis of data or results beyond the assistance with model-output interpretation disclosed above. All AI-assisted content was reviewed and verified by the authors. We take responsibility for the final content of this work, including all text, interpretations, mathematical claims, proofs, explanations, and other artifacts produced with the aid of generative AI.

\subsection*{Reproducibility statement}
Sections~\ref{sec:motif-representation}--\ref{sec:fme} define the model, canonicalization procedure, and explanation identities. Appendix~\ref{app:proofs} provides the proofs and technical derivations. Appendix~\ref{app:experimental_details} reports the datasets, preprocessing, motif-discovery procedure, optimization settings, hyperparameter selection, random seeds, and evaluation protocols. Appendix~\ref{app:additional_experiments} provides the channel ablations.

\bibliography{ref}
\bibliographystyle{iclr2027_conference}

\appendix

\section{Proofs and Technical Details}
\label{app:proofs}

\subsection{Proof of Theorem~\ref{thm:anova}}
\label{app:proof-anova}

All expectations below are taken with respect to the graph-balanced marginals or their product $\mu=\mu_X^{\mathrm{gb}}\otimes\mu_C^{\mathrm{gb}}$. 
Define
\begin{align*}
g_0
&=
\mathbb{E}_{\mu}[g(X,C)],\\
g_X(x)
&=
\mathbb{E}^{\mathrm{gb}}_C[g(x,C)]-g_0,\\
g_C(c)
&=
\mathbb{E}^{\mathrm{gb}}_X[g(X,c)]-g_0,\\
g_{XC}(x,c)
&=
g(x,c)-g_0-g_X(x)-g_C(c).
\end{align*}
The last identity gives
Equation~\ref{eq:canonical-decomposition}. Moreover,
\begin{align*}
\mathbb{E}^{\mathrm{gb}}_X[g_X(X)]
&=
\mathbb{E}_{\mu}[g(X,C)]-g_0
=
0,\\
\mathbb{E}^{\mathrm{gb}}_C[g_C(C)]
&=
\mathbb{E}_{\mu}[g(X,C)]-g_0
=
0.
\end{align*}
For every $c\in\mathcal C_{\mathrm{tr}}$,
\begin{align*}
\mathbb{E}^{\mathrm{gb}}_X[g_{XC}(X,c)]
&=
\mathbb{E}^{\mathrm{gb}}_X[g(X,c)]
-g_0
-\mathbb{E}^{\mathrm{gb}}_X[g_X(X)]
-g_C(c)\\
&=
0.
\end{align*}
The same calculation gives
\[
\mathbb{E}^{\mathrm{gb}}_C[g_{XC}(x,C)]
=
0
\qquad
\text{for every }x\in\mathcal X_{\mathrm{tr}}.
\]

The four components are pairwise orthogonal under
$\langle p,q\rangle_\mu=\mathbb{E}_{\mu}[p(X,C)q(X,C)]$.
The zero-mean identities make each nonconstant component orthogonal to
the constant component. For the remaining pairs,
\begin{align*}
\langle g_X,g_C\rangle_\mu
&=
\mathbb{E}^{\mathrm{gb}}_X[g_X(X)]
\mathbb{E}^{\mathrm{gb}}_C[g_C(C)]
=
0,\\
\langle g_X,g_{XC}\rangle_\mu
&=
\mathbb{E}^{\mathrm{gb}}_X
\left[
g_X(X)
\mathbb{E}^{\mathrm{gb}}_C[g_{XC}(X,C)]
\right]
=
0,\\
\langle g_C,g_{XC}\rangle_\mu
&=
\mathbb{E}^{\mathrm{gb}}_C
\left[
g_C(C)
\mathbb{E}^{\mathrm{gb}}_X[g_{XC}(X,C)]
\right]
=
0.
\end{align*}

For uniqueness, subtract two decompositions satisfying the stated
constraints. Denote the differences by
$\delta_0$, $\delta_X$, $\delta_C$, and $\delta_{XC}$. They are pairwise
orthogonal by the argument above and satisfy
\[
\delta_0+\delta_X(x)+\delta_C(c)+\delta_{XC}(x,c)=0.
\]
Taking the $\mu$-expectation of the square gives
\[
0
=
\delta_0^2
+
\mathbb{E}^{\mathrm{gb}}_X[\delta_X(X)^2]
+
\mathbb{E}^{\mathrm{gb}}_C[\delta_C(C)^2]
+
\mathbb{E}_{\mu}[\delta_{XC}(X,C)^2].
\]
Every term is nonnegative, so every term is zero. Each point in the
empirical support has positive mass. Therefore,
\[
\delta_0=0,\qquad
\delta_X(x)=0,\qquad
\delta_C(c)=0,\qquad
\delta_{XC}(x,c)=0
\]
for every
$x\in\mathcal X_{\mathrm{tr}}$,
$c\in\mathcal C_{\mathrm{tr}}$, and
$(x,c)\in\mathcal X_{\mathrm{tr}}\times\mathcal C_{\mathrm{tr}}$,
respectively. This proves uniqueness.

Finally, consider $H(x,c)=A(x)B(c)$ and define
\[
\overline A
=
\mathbb{E}^{\mathrm{gb}}_X[A(X)],
\qquad
\overline B
=
\mathbb{E}^{\mathrm{gb}}_C[B(C)].
\]
Then
\begin{align*}
A(x)B(c)
&=
\overline A\,\overline B
+
\overline B\bigl(A(x)-\overline A\bigr)
+
\overline A\bigl(B(c)-\overline B\bigr)\\
&\quad+
\bigl(A(x)-\overline A\bigr)
\bigl(B(c)-\overline B\bigr).
\end{align*}
The four terms are, respectively, the constant, feature, structural,
and interaction components. The last term has zero graph-balanced mean
in either argument. By uniqueness,
\[
H_{XC}(x,c)
=
\bigl(A(x)-\overline A\bigr)
\bigl(B(c)-\overline B\bigr).
\]
\hfill\(\square\)

\subsection{Occurrence allocation and permutation invariance}
\label{app:occurrence-allocation}

Let \(c_m^0=\psi_m(0)\) denote the transformed value at zero motif
participation, and define
\[
    \Delta^S_{vm}
    =
    s_m(c_{vm})-s_m(c_m^0),
    \qquad
    \Delta^I_{vkm}
    =
    a_k(x_{vk})
    \left[b_m(c_{vm})-b_m(c_m^0)\right].
\]
For an occurrence \(U\in\mathcal{O}_m(G)\), define
\begin{equation}
    \phi^S_m(U)
    =
    \sum_{v\in U}\frac{\Delta^S_{vm}}{h_{vm}},
    \qquad
    \phi^I_{km}(U)
    =
    \sum_{v\in U}\frac{\Delta^I_{vkm}}{h_{vm}}.
    \label{eq:occurrence-s-i}
\end{equation}
Since node \(v\) belongs to exactly \(h_{vm}\) occurrences of motif
\(m\),
\begin{align*}
    \sum_{U\in\mathcal{O}_m(G)}\phi^S_m(U)
    &=
    \sum_{v:h_{vm}>0}
    \frac{\Delta^S_{vm}}{h_{vm}}
    \left|\{U\in\mathcal{O}_m(G):v\in U\}\right|\\
    &=
    \sum_{v:h_{vm}>0}\Delta^S_{vm}
    =
    \sum_{v\in V}\Delta^S_{vm},
\end{align*}
where the final equality follows because \(h_{vm}=0\) implies
\(\Delta^S_{vm}=0\). The same argument gives
\[
    \sum_{U\in\mathcal{O}_m(G)}\phi^I_{km}(U)
    =
    \sum_{v\in V}\Delta^I_{vkm}.
\]
Adding the zero-participation terms therefore yields
\begin{align*}
    Q^S_m(G)
    &=
    \sum_{v\in V}s_m(c_m^0)
    +
    \sum_{U\in\mathcal{O}_m(G)}\phi^S_m(U),\\
    Q^I_{km}(G)
    &=
    \sum_{v\in V}a_k(x_{vk})b_m(c_m^0)
    +
    \sum_{U\in\mathcal{O}_m(G)}\phi^I_{km}(U).
\end{align*}
This argument applies separately to each motif type and is unaffected by
overlapping occurrences. Finally, under any vertex permutation
\(\pi:V\to V\), the map \(U\mapsto\pi(U)\) is a bijection between
corresponding motif occurrences and preserves \(x_v\), \(h_{vm}\), and
\(c_{vm}\). Hence graph-level contributions are invariant, while node
and occurrence explanations are transported equivariantly.

\subsection{Sparsity Penalty and Reciprocal-Scale Balancing}
\label{app:sparsity-scale}

The sparsity penalty used in the training objective is
\begin{equation}
\Omega =
\sum_{k=1}^{d}
\mathbb{E}^{\mathrm{gb}}_X
\left[
    |f_k(X_k)| + |a_k(X_k)|
\right]
+
\sum_{m=1}^{M}
\mathbb{E}^{\mathrm{gb}}_C
\left[
    |s_m(C_m)| + |b_m(C_m)|
\right].
\end{equation}
The terms involving $f_k$ and $s_m$ promote sparse main-effect
explanations. The terms involving $a_k$ and $b_m$ also select a magnitude
within the reciprocal-rescaling class of the factorized interaction.

Define
\[
R_A =
\sum_{k=1}^{d}
\mathbb{E}^{\mathrm{gb}}_X |a_k(X_k)|,
\qquad
R_B =
\sum_{m=1}^{M}
\mathbb{E}^{\mathrm{gb}}_C |b_m(C_m)|.
\]
Under $a_k\mapsto\gamma a_k$ and
$b_m\mapsto b_m/\gamma$, the interaction remains unchanged, while its
contribution to the penalty becomes
\[
R_I(\gamma)
=
|\gamma|R_A + \frac{R_B}{|\gamma|}.
\]
For $R_A,R_B>0$, the arithmetic--geometric mean inequality gives
\[
R_I(\gamma)\geq 2\sqrt{R_A R_B},
\]
with equality when
\[
|\gamma|=\sqrt{\frac{R_B}{R_A}}.
\]
Thus, among equivalent factorizations, the penalty selects the magnitude
that equalizes the two penalized factors. The global sign remains
unidentified, but it cancels in every product
$a_k(x_k)b_m(c_m)$ and therefore does not affect any reported
feature--motif contribution. If either $R_A$ or $R_B$ is zero, the
interaction vanishes almost everywhere under the reference distribution,
and its scale is immaterial.

\section{Experimental Details}
\label{app:experimental_details}

\subsection{Datasets and Preprocessing}
\label{app:datasets}

We use Mutagenicity, NCI1, PROTEINS, and PTC-MR from
TUDataset~\cite{morris2020tudataset} for binary classification, loaded
with PyTorch Geometric 2.7.0. Regression benchmarks comprise BA-Reg~1,
BA-Reg~2, and Crippen from \cite{zhang2024regexplainer}, together with the
$\mu$, $\alpha$, and HOMO targets from QM9~\cite{ramakrishnan2014quantum}.
BA-Reg~1 targets the sum of node features inside planted house motifs;
BA-Reg~2 targets the number of planted houses. Crippen targets the
dimensionless Wildman--Crippen $\log P$. QM9 reports $\mu$ in Debye,
$\alpha$ in $a_0^3$, and HOMO energy in eV. Label~1 defines the positive
class; for Mutagenicity, it denotes non-mutagenic compounds.

We remove self-loops and represent each input as a simple undirected graph.
For \MI, each one-hot group is treated as one categorical variable and
mapped through a learned lookup table. Numerical features are standardized
coordinate-wise using training-node statistics, with unit scale for
constant coordinates. Regression targets are standardized on the training
graphs and returned to their original units before computing MAE. All
statistics, motif discovery, and motif normalization are fitted on the
training split and then frozen.

We use a 70/15/15 train/validation/test split for each of five seeds, $\{11,22,33,44,55\}$. Each seed determines both the data split and model training. The three QM9 tasks share the same graph indices within each seed. Reported results are averaged over these five seeds, with preprocessing fitted independently on each training split.

\begin{table*}[ht]
\centering
\setlength{\tabcolsep}{4pt}
\caption{Dataset statistics. B and R denote binary classification and
regression. Edges count undirected pairs; feature dimensions are shown
before and after \MI encoding.}
\label{tab:datasets}
\begin{tabular}{lclrrc}
\toprule
Dataset & Task & Train/Val/Test & Avg.\ nodes & Avg.\ edges & Features \\
\midrule
Mutagenicity
& B & 3035/651/651 & 30.32 & 30.77 & $14\rightarrow1$ \\
NCI1
& B & 2877/616/617 & 29.87 & 32.30 & $37\rightarrow1$ \\
PROTEINS
& B & 779/167/167 & 39.06 & 72.82 & $4\rightarrow2$ \\
PTC-MR
& B & 240/52/52 & 14.29 & 14.69 & $18\rightarrow1$ \\
BA-Reg 1
& R & 700/150/150 & 25.00 & 25.00 & $10\rightarrow10$ \\
BA-Reg 2
& R & 700/150/150 & 120.00 & 128.90 & $10\rightarrow10$ \\
Crippen
& R & 788/169/170 & 25.66 & 26.04 & $14\rightarrow1$ \\
QM9 ($\mu,\alpha,\mathrm{HOMO}$)
& R & 91581/19625/19625 & 18.03 & 18.66 & $11\rightarrow7$ \\
\bottomrule
\end{tabular}
\end{table*}

\subsection{Motif Discovery Protocol}
\label{sec:motif-discovery}

For a candidate graphlet $P$, define
\[
\mathcal{O}_P(G)
=
\{U\subseteq V: |U|=q,\;G[U]\simeq P\},
\qquad
N_P(G)=|\mathcal{O}_P(G)|,
\]
and let
\[
C_P=\sum_{G\in\mathcal{D}_{\mathrm{tr}}}N_P(G)
\]
be its total training-set count. If $C_P^{(\ell)}$ denotes the corresponding
count in the $\ell$-th degree-preserving randomized training set, its
one-sided empirical $p$-value is
\[
p_P
=
\frac{
1+\sum_{\ell=1}^{L}
\mathbf{1}[C_P^{(\ell)}\geq C_P]
}{
L+1
}.
\]
Candidates are selected when their Benjamini--Hochberg adjusted $p$-value
is at most $\alpha$ and their observed count exceeds the null mean.

We consider all 29 connected, non-isomorphic unlabeled graphlets with $q\in\{3,4,5\}$. Their induced occurrences are enumerated exactly over each seed-specific training split using \texttt{igraph}'s ESU  implementation~\citep{wernicke2006fanmod}.

Each null replica independently rewires every training graph from its original state using simple-graph double-edge swaps, with $60|E(G)|$ attempts. Degrees are preserved, while self-loops and parallel edges are forbidden. We generate $L=500$ replicas with seed 31415. A candidate is selected when its Benjamini--Hochberg adjusted enrichment $p$-value is at most $0.05$ and its observed count exceeds the null mean. No vocabulary cap is imposed. We repeat motif discovery independently for each evaluation seed. For seed \(r\), candidate counts, degree-preserving null graphs, and motif selection are computed using only the corresponding training split \(D_{\mathrm{tr}}^{(r)}\). The resulting vocabulary \(\mathcal{P}^{(r)}\) and motif-normalization statistics are frozen and applied to the validation and test graphs of the same split. No validation or test graph is used for motif discovery or normalization.

For each selected motif, the mean and population standard deviation of $\log(1+h_{vm})$ are computed over all training nodes. Standard deviations below $10^{-8}$ are set to one. The vocabulary and normalization statistics are then frozen. Synthetic effect recovery bypasses discovery and uses a fixed vocabulary containing the three-node path and triangle.

\begin{table}[t]
\centering
\small
\caption{Selected motif-vocabulary sizes across the five seed-specific training splits. The three QM9 targets share the same seed-specific vocabulary because they use identical graph indices within each seed.}
\label{tab:motif-vocabularies}
\begin{tabular}{lr@{\qquad}lr}
\toprule
Dataset & Selected & Dataset & Selected \\
\midrule
Mutagenicity & 4  & BA-Reg 1         & 7 \\
NCI1         & 7  & BA-Reg 2         & 5 \\
PROTEINS     & 21 & Crippen          & 4 \\
PTC-MR       & 6  & QM9 (all targets)& 7 \\
\bottomrule
\end{tabular}
\end{table}

We additionally measure the wall-clock time required for this exact enumeration over the complete training split. As reported in Table~\ref{tab:motif_runtime}, enumerating all 29 candidate graphlets takes less than 0.6 seconds on every dataset except QM9, where all 91,581 training graphs are processed in 2.52 seconds.

\begin{table}[t]
\centering
\caption{
Wall-clock time required to enumerate all 29 candidate graphlets, averaged over the five seed-specific training splits. The QM9 time is averaged over the three equivalent measurements.}
\label{tab:motif_runtime}
\begin{tabular}{lrr}
\toprule
Dataset & Selected motifs & Enumeration time (s) \\
\midrule
Mutagenicity      & 4  & 0.148 \\
NCI1              & 7  & 0.128 \\
PROTEINS          & 21 & 0.164 \\
PTC-MR            & 6  & 0.005 \\
BA-Reg 1          & 7  & 0.044 \\
BA-Reg 2          & 5  & 0.580 \\
Crippen           & 4  & 0.028 \\
QM9 (all targets) & 7  & 2.523 \\
\bottomrule
\end{tabular}
\end{table}

\subsection{Training and Model Selection}
\label{app:training}

All models are trained with AdamW for at most 500 epochs, with early-stopping patience 50. We use binary cross-entropy with logits for classification and mean squared error on standardized targets for regression. Graphs are shuffled at each epoch, and gradients are clipped to a global $\ell_2$ norm of five. For \MI, graph-balanced centering statistics are refreshed before training and after each epoch.

Checkpoints are selected using the validation objective employed during training. Improvements smaller than \(10^{-6}\) are ignored, and the best checkpoint is restored for test evaluation. Final regression performance is reported as MAE in the original target scale.

For each dataset, we randomly sample and evaluate 30 hyperparameter configurations for \MI, GNAN, GCN, GIN, and Graph Transformer. Configurations are sampled from the search space defined by depth \(\{2,3\}\), hidden dimension \(\{16,32,64\}\), dropout \(\{0,0.05,0.10\}\), learning rate \(\eta \in \{10^{-3},10^{-2},10^{-1}\}\), weight decay \(\lambda_{\mathrm{wd}} \in \{10^{-5},10^{-4}\}\), and batch size \(\{128,256,512\}\). For \MI, the search space additionally includes the sparsity coefficient \(\lambda_{\mathrm{sp}} \in \{10^{-3},10^{-2},10^{-1}\}\)

For G-NAMRFF, we apply the same random-search protocol, independently sampling and evaluating 30 configurations for each dataset. Its search space contains \(M \in \{20,40,50,100,200\}\) random Fourier features, \(R \in \{1,\ldots,7\}\) graph-filter steps, batch size \(\{128,256,512\}\), and learning rate \(\eta \in \{10^{-3},10^{-2},10^{-1}\}\). The remaining parameters are sampled as

\[
\theta\sim\operatorname{LogUniform}(1,4),\qquad
\lambda_{\mathrm{wd}}\sim
\operatorname{LogUniform}(10^{-5},10^{-2}).
\]

For every dataset and model family, all 30 randomly sampled configurations are evaluated under the training protocol described above and ranked by their mean best-validation score. The selected configuration is then retrained from scratch for each of the five evaluation seeds listed in Section~\ref{app:datasets}, using the train/validation/test split generated by the corresponding seed.

\paragraph{Training efficiency.}
Table~\ref{tab:runtime} reports the wall-clock training time of the selected
configurations, averaged over the five final runs. \MI has the lowest mean
runtime on nine of ten datasets; Graph Transformer is faster only on
BA-Reg~2. The difference is largest on QM9, where \MI trains roughly one
order of magnitude faster than the baselines. All models were evaluated
under the same hardware conditions. These measurements cover final model
training and exclude hyperparameter search and the one-time motif-discovery
step.
\begin{table}[t]
\small
\centering
\setlength{\tabcolsep}{2.8pt}
\caption{Wall-clock training time in seconds, averaged over the five final
runs and rounded to the nearest second. Hyperparameter search and motif
discovery are excluded. Lower is better; the best result is bold.}
\label{tab:runtime}
\begin{tabular}{lrrrrrr}
\toprule
Dataset & GCN & GIN & GTrans & GNAN & G-NAMRFF & \MI \\
\midrule
Mutagenicity
& $108 \pm 25$ & $60 \pm 22$ & $51 \pm 8$
& $193 \pm 61$ & $838 \pm 443$ & $\mathbf{19 \pm 9}$ \\

NCI1
& $99 \pm 35$ & $93 \pm 23$ & $81 \pm 19$
& $132 \pm 46$ & $958 \pm 440$ & $\mathbf{14 \pm 3}$ \\

PROTEINS
& $21 \pm 9$ & $16 \pm 7$ & $19 \pm 7$
& $60 \pm 9$ & $125 \pm 41$ & $\mathbf{7 \pm 1}$ \\

PTC-MR
& $2 \pm 0$ & $3 \pm 1$ & $4 \pm 1$
& $3 \pm 1$ & $8 \pm 3$ & $\mathbf{2 \pm 1}$ \\

BA-Reg 1
& $43 \pm 2$ & $38 \pm 8$ & $28 \pm 9$
& $100 \pm 29$ & $119 \pm 30$ & $\mathbf{25 \pm 8}$ \\

BA-Reg 2
& $97 \pm 0$ & $75 \pm 23$ & $\mathbf{37 \pm 15}$
& $187 \pm 136$ & $115 \pm 41$ & $56 \pm 8$ \\

Crippen
& $45 \pm 4$ & $45 \pm 5$ & $41 \pm 9$
& $86 \pm 35$ & $92 \pm 0$ & $\mathbf{6 \pm 1}$ \\

QM9-$\alpha$
& $2978 \pm 4$ & $2688 \pm 19$ & $2856 \pm 34$
& $4512 \pm 2175$ & $2409 \pm 859$ & $\mathbf{240 \pm 10}$ \\

QM9-HOMO
& $2526 \pm 4$ & $2467 \pm 6$ & $2743 \pm 40$
& $6554 \pm 1272$ & $4463 \pm 493$ & $\mathbf{304 \pm 103}$ \\

QM9-$\mu$
& $2619 \pm 146$ & $2595 \pm 19$ & $2839 \pm 154$
& $4718 \pm 1773$ & $3869 \pm 1022$ & $\mathbf{241 \pm 9}$ \\
\bottomrule
\end{tabular}
\end{table}

\subsection{Explanation Evaluation Protocol}
\label{app:explanation_evaluation}

We evaluate node-level plausibility on every test graph. Ground-truth positive nodes are the five vertices of the planted motif in BA-Reg~1 and the union of all planted-motif vertices in BA-Reg~2; all remaining nodes are negative. 

For \MI, nodes are ranked by the magnitude of their native contribution,
\[
s_v^{\MI}=|F_v+S_v+I_v|,
\]
without allocating the graph intercept. For GNAN~\cite{bechler2024intelligible} and G-NAMRFF~\cite{siddartha2026interpretable}, we use the node-level relevance scores defined by their respective papers, following the original aggregation, baseline, and target-rescaling conventions. 

GNNExplainer and Integrated Gradients explain the corresponding trained GIN. GNNExplainer uses PyG~2.7.0 with one object mask per node and no edge mask. The mask is optimized for 200 Adam steps with learning rate $0.01$, using size and entropy penalties $1.0$ and $0.1$. Integrated Gradients uses an all-zero raw-feature baseline and 128 trapezoidal integration intervals. Signed feature attributions are summed per node before taking their absolute value.

For each seed, AUROC is computed separately for every graph in the corresponding test split and then averaged across test graphs with equal weight, yielding one aggregate score per seed. Table~\ref{tab:bareg-auroc} reports the mean and sample standard deviation of these aggregate scores across the five training seeds \(\{11,22,33,44,55\}\).

\subsection{Synthetic Recovery Details}
\label{app:synthetic_details}

For each of the five seeds $\{11, 22, 33, 44, 55\}$, an independent 1,400/300/300 training/validation/test split is generated and the model is trained from scratch, with checkpoint selection based on validation RMSE.

The five runs use training seeds $\{11,22,33,44,55\}$. Observation noise
is generated independently with seed 8193 and standard deviation
$\sigma=1.2312$, equal to 10\% of the population standard deviation of
the clean training targets.

The experiment uses a fixed \MI configuration with $1$--$32$--$1$ scalar
MLPs, ReLU activations, and no dropout. Training uses AdamW with learning
rate $3\times10^{-3}$, weight decay $10^{-5}$, sparsity coefficient
$10^{-4}$, batch size 128, and gradient clipping at five. Models train for
at most 600 epochs with patience 100, and the checkpoint with the lowest
validation MSE is restored.

For evaluation, the learned graph-balanced components are re-anchored at
raw feature value $x=0$ and raw motif count $h=0$. Input preprocessing is
included in the shape-function arguments. Define
\[
\Delta a_j(x)=a_j(x)-a_j(0),\qquad
\Delta b_k(h)=b_k(h)-b_k(0),
\]
and
\[
A_0=\sum_j a_j(0),\qquad B_0=\sum_k b_k(0),\qquad
F_0=\sum_j f_j(0),\qquad S_0=\sum_k s_k(0).
\]
Writing $\mu_y$ and $\sigma_y$ for the training-target mean and scale, the
zero-reference components in original target units are
\begin{align*}
f_j^{(0)}(x)
&=\sigma_y\!\left[f_j(x)-f_j(0)+B_0\Delta a_j(x)\right],\\
s_k^{(0)}(h)
&=\sigma_y\!\left[s_k(h)-s_k(0)+A_0\Delta b_k(h)\right],\\
I_{jk}^{(0)}(x,h)
&=\sigma_y\,\Delta a_j(x)\Delta b_k(h).
\end{align*}
Since every graph has 12 nodes, the corresponding intercept is
\[
c^{(0)}
=\mu_y+\sigma_y\!\left[c+12\left(F_0+S_0+A_0B_0\right)\right].
\]
This transformation preserves predictions exactly. Node-level feature,
structural, and interaction channels are obtained by summing their
respective re-anchored components.

The clean ground-truth channels are
\[
F_v^\star=3\sin(x_{v1}),\qquad
S_v^\star=3(1-e^{-h_{v1}/5}),\qquad
I_v^\star=3\tanh(x_{v2})(1-e^{-h_{v2}/2}).
\]
Noise is applied only to the graph target. Channel recovery is measured by
RMSE over all test nodes, while graph-level accuracy is measured by MAE
against the noisy targets.

Figure~\ref{fig:toy-recovery} uses training seed 11. Its local panels show
the first test graph containing both motif types, selected without reference
to prediction or recovery error; the recovery panel includes every test
node.

\section{Additional Experiments}
\label{app:additional_experiments}

Table~\ref{tab:ablation} isolates the feature ($F$), structural ($S$),
and feature--motif interaction ($I$) channels. The full model obtains
the best mean performance on nine of the ten datasets. The two synthetic
tasks expose the distinct roles of structure and interaction. BA-Reg 1
depends on node features within planted motifs: adding $I$ reduces MAE
from $449.50$ to $10.44$, while combining all channels further reduces it
to $4.13$. In contrast, BA-Reg 2 depends directly on the number of planted
motifs: adding $S$ reduces MAE from $5.1492$ to $0.0097$, whereas $F+I$
does not improve over the feature-only model. On the real-world
benchmarks, the relative benefits of $S$ and $I$ are task-dependent, but
their combination generally provides the strongest performance.
NCI1 is the only exception, where $F+I$ marginally outperforms the
full model.

\begin{table}[ht]
\centering
\small
\begin{tabular}{@{}lcccc@{}}
\toprule
Dataset & $F$ & $F+S$ & $F+I$ & $F+S+I$ \\
\midrule
\multicolumn{5}{l}{\textit{Classification: test AUROC (\%), $\uparrow$}} \\
\midrule
Mutagenicity
& $75.48 \pm 0.06$
& $80.00 \pm 0.37$
& $80.09 \pm 0.26$
& $\mathbf{82.81 \pm 1.50}$ \\

NCI1
& $64.24 \pm 0.46$
& $75.76 \pm 0.29$
& $\mathbf{78.17 \pm 0.68}$
& $76.81 \pm 2.01$ \\

PROTEINS
& $64.90 \pm 0.95$
& $70.36 \pm 0.84$
& $65.44 \pm 1.08$
& $\mathbf{73.32 \pm 4.63}$ \\

PTC-MR
& $49.69 \pm 3.99$
& $54.09 \pm 3.74$
& $49.69 \pm 3.99$
& $\mathbf{57.43 \pm 2.62}$ \\
\midrule
\multicolumn{5}{l}{\textit{Regression: test MAE, $\downarrow$}} \\
\midrule
BA-Reg 1
& $449.50 \pm 1.90$
& $462.40 \pm 7.05$
& $10.44 \pm 11.36$
& $\mathbf{4.2 \pm 1.2}$ \\

BA-Reg 2
& $5.1492 \pm 0.0010$
& $0.0097 \pm 0.0101$
& $5.1501 \pm 0.0012$
& $\mathbf{0.0087 \pm 0.0070}$ \\

Crippen
& $0.938 \pm 0.015$
& $0.835 \pm 0.033$
& $0.612 \pm 0.028$
& $\mathbf{0.460 \pm 0.181}$ \\

QM9-$\mu$
& $0.9785 \pm 0.0006$
& $0.8936 \pm 0.0011$
& $0.8639 \pm 0.0038$
& $\mathbf{0.8519 \pm 0.0121}$ \\

QM9-$\alpha$
& $3.5918 \pm 0.0221$
& $3.3515 \pm 0.0331$
& $3.5917 \pm 0.0221$
& $\mathbf{3.3510 \pm 0.0315}$ \\

QM9-HOMO
& $0.4158 \pm 0.0013$
& $0.2813 \pm 0.0010$
& $0.2910 \pm 0.0211$
& $\mathbf{0.2402 \pm 0.0131}$ \\
\bottomrule
\end{tabular}
\caption{
Channel ablation across five independently generated test splits. Entries report the mean $\pm$ standard deviation over five seeds, each defining both the data split and model training.
The $F+S+I$ column reuses the full-model runs reported in the main results. Bold indicates the best mean in each row.
}
\label{tab:ablation}
\end{table}

\end{document}